\documentclass[letterpaper, 10 pt, conference]{ieeeconf}

\usepackage{graphicx}
\usepackage{amsmath}
\usepackage{amssymb}
\usepackage{booktabs}
\usepackage{multirow}
\usepackage{subfigure}
\usepackage{epstopdf}

\usepackage[dvipsnames,table,xcdraw]{xcolor}
\usepackage[ruled,vlined,linesnumbered, boxed]{algorithm2e}

\usepackage{yhmath}
\usepackage{cite}

\newtheorem{conclusion}{\bf Conclusion}
\newtheorem{theorem}{\bf Theorem}
\DeclareMathOperator{\tr}{Tr}
\DeclareMathOperator{\diag}{diag}
\DeclareMathOperator{\erf}{erf}
\makeatletter
\let\NAT@parse\undefined
\makeatother
\usepackage[backref, colorlinks, linkcolor=red, anchorcolor=blue, citecolor=blue]{hyperref}

\usepackage{listings} % File listings, with syntax highlighting
\usepackage{bm}

\usepackage[framemethod=tikz]{mdframed} % Allows defining custom boxed/framed environments
\usepackage{perpage} % The perpage package allows resetting of counters per page
\MakePerPage{myfootnote} % Reset the footnote counter per page

\mdfdefinestyle{file}{
	innertopmargin=0.8\baselineskip,
	innerbottommargin=0.1\baselineskip,
	topline=false, bottomline=false,
	leftline=false, rightline=false,
	leftmargin=0.0cm,
	rightmargin=0cm,
	singleextra={%
		\draw[fill=black!10!white](P)++(0,-1.2em)rectangle(P-|O);
		\node[anchor=north west]
		at(P-|O){\rmfamily\mdfilename};
		\def\l{0.5em}
		\draw(O-|P)++(-\l,0)--++(\l,\l)--(P)--(P-|O)--(O)--cycle;
		\draw(O-|P)++(-\l,0)--++(0,\l)--++(\l,0);
	},
	nobreak,
}

\vspace{0.8cm}  %调整图片与上文的垂直距离

\IEEEoverridecommandlockouts                              % This command is only needed if 
\title{\LARGE \bf
Tight Collision Probability for UAV Motion Planning \\ in Uncertain Environment
}
\author{Tianyu Liu$^{1}$, Fu Zhang$ ^{1} $, Fei Gao$^{2}$,  and Jia Pan$^{3} $
	\thanks{1 Department of Mechanical Engineering, The University of Hong Kong.} 
	\thanks{2 Institute of Cyber-Systems and Control, Zhejiang University.}
	\thanks{3 Department of Computer Science, The University of Hong Kong.}
	\thanks{Corresponding author: Tianyu Liu, {\tt\small tianyu@connect.hku.hk}.}
}

\begin{document}

\maketitle
\thispagestyle{empty}
\pagestyle{empty}

%%%%%%%%%%%%%%%%%%%%%%%%%%%%%%%%%%%%%%%%%%%%%%%%%%%%%%%%%%%%%%%%%%%%%%%%%%%%%%%%
\begin{abstract}
Operating unmanned aerial vehicles (UAVs) in complex environments that feature dynamic obstacles and external disturbances poses significant challenges, primarily due to the inherent uncertainty in such scenarios. Additionally, inaccurate robot localization and modeling errors further exacerbate these challenges. Recent research on UAV motion planning in static environments has been unable to cope with the rapidly changing surroundings, resulting in trajectories that may not be feasible. Moreover, previous approaches that have addressed dynamic obstacles or external disturbances in isolation are insufficient to handle the complexities of such environments. This paper proposes a reliable motion planning framework for UAVs, integrating various uncertainties into a \textit{chance constraint} that characterizes the uncertainty in a probabilistic manner. The chance constraint provides a probabilistic safety certificate by calculating the collision probability between the robot's Gaussian-distributed forward reachable set and states of obstacles. To reduce the conservatism of the planned trajectory, we propose a tight upper bound of the collision probability and evaluate it both exactly and approximately. The approximated solution is used to generate motion primitives as a reference trajectory, while the exact solution is leveraged to iteratively optimize the trajectory for better results. Our method is thoroughly tested in simulation and real-world experiments, verifying its reliability and effectiveness in uncertain environments.
\end{abstract}

\section{Introduction}
%As the development of the safe motion planning technology, UAVs are increasingly deployed in the industry for various tasks such as inspection and delivery. 
Nowadays, UAVs are increasingly deployed in various scenarios and are facing more complicated operating environments. They may fly under unknown air turbulence in the wild or work in close physical proximity to walking humans in the logistics factory. These uncertain environments, along with robot intrinsic uncertainty such as modeling errors and inaccurate localization, makes safe motion planning a challenging and computationally expensive problem \cite{blackmore2011chance, majumdar2017funnel}. Online planners designed for deterministic environments are inadequate to handle uncertainty, relying solely on replanning mechanisms or feedback controllers. Thus, it is essential for motion planners to reason about uncertainty online and react intelligently with safety guarantees.
\begin{figure}
%	\vspace{0.2cm}
	\centering
	\includegraphics[width=1.0\linewidth]{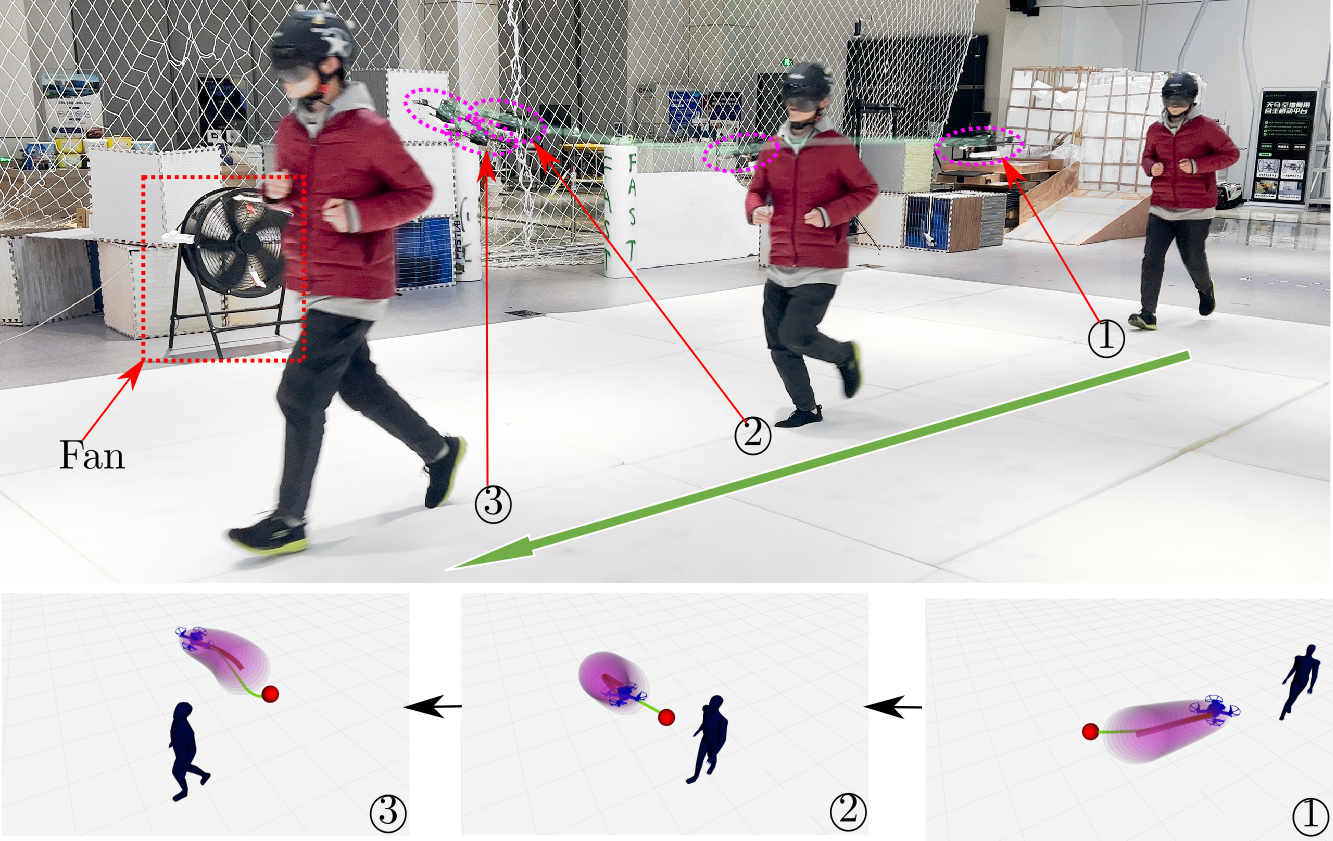}
	\caption{The UAV aims to hover at a fixed point in a windy environment while actively avoiding a moving human. \textbf{Top}: snapshots of real world experiment. \textbf{Bottom}: data visualization in Rviz. \textbf{Red ball}: goal point.  \textbf{Green line}: reference trajectory. \textbf{Brown line}: optimized trajectory. \textbf{Purple ellipsoid}: FRS.}
	\label{headpic}
	\vspace{-0.5cm}
\end{figure}

%can be too conservative as the probabilities are usually loose upper bounds.

The majority of prior research on planning under uncertainty can be divided into \textit{chance constrained approach}, and \textit{set bounded approach}. The former leverages a probabilistic representation of uncertainty and guarantees the probability of failure is below a specified threshold \cite{blackmore2011chance}. This approach has been adopted to handle uncertainty arising from dynamic obstacles \cite{zhu2019chance, lin2020robust, castillo2020real} and system modeling errors \cite{dai2019chance}. The latter method typically ensures safety certification for robots by achieving robust planning \cite{manchester2017dirtrel} while assuming a disturbance bound, particularly for wind \cite{seo2019robust, majumdar2017funnel, manchester2017dirtrel}. In this paper, we present a reliable motion planning framework for UAVs that combines both approaches to address the uncertainties associated with dynamic obstacles and wind disturbances. Specifically, we propose an integrated chance constraint that restricts the collision probability between the robot's Gaussian-distributed forward reachable set (FRS) and the predicted states of obstacles. However, no closed-form solution exists for estimating this probability, and previous methods have either evaluated loose upper bounds or approximations with significant errors.  With various kinds of uncertainty considered, these collision estimation methods will result in overly conservative and inefficient trajectories. To this end, we propose a tight probability upper bound and give its exact and approximated solutions. The approximated solution is computed from numerical integral using Gauss-Hermit quadrature. It is served for collision chance checking of motion primitives, which compose the reference trajectory for optimization. 
The exact solution is derived from the series expansions of a quadratic form in random Gaussian variables. It is utilized to iteratively optimize the trajectory for better results. As shown in \hyperref[headpic]{Figure \ref{headpic}}, the real-world experiment demonstrated the reliability and effectiveness of our planning framework.

To conclude, the contributions of this paper are summarized as follows:
\begin{itemize}
	\item A tight collision probability upper bound with its exact and approximated solutions.
	\item A reliable motion planning framework for UAVs in highly uncertain environments with dynamic obstacles and external disturbances.
	\item Extensive Validation of the proposed method in simulation and real-world experiments.
\end{itemize}
%The rest of the paper is organized as follows: Section \ref{lr} reviews existing works. Section \ref{ps} describes the formalized motion planning framework. Section \ref{tcc} presents our approaches for collision probability evaluation, the iterative method for trajectory optimization, and the corresponding benchmark comparisons. Section \ref{ump} integrates the proposed methods into our UAV planning framework and presents the benchmark comparisons in simulation and validation in real-world experiments. Finally, Section \ref{cln} concludes the paper and outlines future work.

% The methods are further integrated into our UAV planning framework, which is detailed in Section \ref{ump}. The benchmark comparisons in simulation and validation in real-world experiments are presented in this section too. Conclusions and future work is stated in Section \ref{cln}.

\section{Literature Review}
\label{lr}
\subsection{Collision Probability Estimation}
Collision probability estimation is essential for motion planning in probabilistic contexts. Monte Carlo (MC) provides a powerful tool for this purpose, but it requires a large number of simulation rollouts. The simulation was  accelerated in \cite{lambert2008fast}  by approximating the double summation from the MC integration with a single summation. Assuming that obstacles are punctual objects,  a representative density point on the distribution can be selected and multiplied by the volume of the robot. Park et al. \cite{park2012itomp} took the maximum density value at the robot surface and provided an upper bound, while Du et al. \cite{du2011probabilistic} took the density at the center. Despite their efficiency, these methods usually yield large errors.  Some other methods marginalized the probability over the collision region of the configuration space.  Hardy et al. \cite{hardy2013contingency}, and Zhu et al. \cite{zhu2019chance} both obtained upper bounds with collision regions as a rectangular bounding box and half hyperspace respectively. Dai et al. \cite{dai2019chance} proposed to approximate the probability via quadrature rules under the assumption that each distribution dimension is independent. Antony et al. \cite{thomas2021exact} claimed to find the exact solution for ellipsoidal-shaped robots and obstacles, but there seems to be an oversight of taking random variables as constant in their theory derivation\footnote{See our comments \href{https://github.com/Acmece/IROS2023_Com}{https://github.com/Acmece/IROS2023\_Com} for more details}. 
%In this work, we propose a tight upper bound and find its exact solution, so that the over-estimated part is only 
 
\subsection{{Planning Under Uncertainty}}
Planing under uncertainty can mainly be divided into \textit{chance constrained approach} \cite{blackmore2011chance, zhu2019chance,castillo2020real,dai2019chance} and \textit{set bounded approach} \cite{seo2019robust, majumdar2017funnel, manchester2017dirtrel, wu2021external, kim2018computing}. Considering the obstacles as convex polyhedrons,  Blackmore et al. \cite{blackmore2011chance} approximated the chance-constrained optimization problem as a disjunctive convex programming. Zhu et al. \cite{zhu2019chance} proposed transforming the chance constraint to a deterministic one, typically for ellipsoidal obstacles. Benefiting from the two strategies, a hybrid solution is proposed in \cite{castillo2020real} to trade off performance and efficiency. These methods are mainly designed to handle uncertain dynamic obstacles. While in \cite{dai2019chance}, the robot motion and state uncertainty is focused, and a chance-constrained motion planning system is proposed for high-dimensional robots. \textit{Set bounded approach} reasons about the influence under bounded disturbances (e.g. air turbulence) and usually provides a safety guarantee. Anirudha et al. \cite{majumdar2017funnel} computed funnel library offline via sum-of-squares programming and sequentially composed them online for robust trajectories. The approach has been extended to multirotor systems, where the rotor drag is explicitly estimated rather than treated as an unknown disturbance \cite{kim2018computing}. Using Hamilton-Jacobi reachability, Seo et al. \cite{seo2019robust} conservatively approximate the FRS with ellipsoidal parameterization, which can be computed analytically online and was further integrated with external force estimator for the quadrotor in \cite{wu2021external}. Manchester et al. \cite{manchester2017dirtrel} computed the ellipsoid bounds around nominal trajectory locally in the closed form under the LQR feedback controller and designed a robustness cost for trajectory optimization.
Apart from these two main categories, the chance constraint can be converted to a set-bounded one by enlarging the robot with 3-$\sigma $ uncertainty ellipsoids \cite{park2012itomp, kamel2017robust}. Wang et al. \cite{wang2021autonomous} conducted dynamic avoidance in font-end path searching and then refine the trajectory by gradient descent. In \cite{ho2021adaptive}, the safety margin library is computed via MC simulations offline and adaptively adjusted on-the-fly for disturbances.

\section{Motion Planning Framework}
\label{ps}
Considering a robot moving in an uncertainty environment with dynamic obstacles, external disturbances as well as static structures, we aim to find efficient trajectories with probabilistic safety certificates.
Given the models of robot and $ M $ dynamic obstacles,
we discretize the  planned trajectory $ (\mathbf{x} \in \mathbb{R}^{n_x}, \mathbf{u} \in \mathbb{R}^{n_u}) $ into $N$ knot points in time. The motion planning problem is then formulated by the following nonlinear programming:
\begin{align}
\min _{\mathbf{u}_0, \ldots, \mathbf{u}_{N-1}} & J\left(\mathbf{u}_0, \ldots, \mathbf{u}_{N-1}, \mathbf{x}_0, \ldots, \mathbf{x}_N\right), & \label{trajopt} \\
\text { s.t. } & \mathbf{x}_{t+1}=f \left(\mathbf{x}_t, \mathbf{u}_t\right) + \mathbf{w}_t, &  \tag{\ref{trajopt}{a}} \label{trajopt_robot}  \\
& \mathbf{y}_{t+1}^i=g^i\left(\mathbf{y}_t^i\right)+ \mathbf{v}_t^i, & \tag{\ref{trajopt}{b}} \label{trajopt_obs}  \\
& \mathcal{E}_t = h(\mathbf{x}_t, \mathbf{u}_t, t), &  \tag{\ref{trajopt}{c}} \label{trajopt_frs}  \\
& \mathbf{w}_t \sim \mathcal{N}\left(\mathbf{0}, W_t\right), \quad \mathbf{v}_t^i \sim \mathcal{N}\left(\mathbf{0}, V_t^i\right), & \tag{\ref{trajopt}{d}} \label{trajopt_noise}  \\
& \mathbf{x}_0 \sim \mathcal{N}\left(\hat{\mathbf{x}}_0, \Sigma_{\mathbf{x}, 0}\right), \quad \mathbf{y}_0 \sim \mathcal{N}\left(\hat{\mathbf{y}}_0^i, \Sigma_{\mathbf{y}, 0}\right), \nonumber \\
& \mathbf{x}_{t+1} \in \mathcal{F}_t, \; \mathbf{u}_t \in \mathcal{U} &    \tag{\ref{trajopt}{e}} \label{trajopt_feas}  \\
& \boldsymbol{\mathcal{P} \Big( \mathcal{G}_t \big(\mathcal{E}_t, y_t^1, \cdots y_t^M \big) < 0 \Big) < \Delta_t}, & \tag{\ref{trajopt}{f}} \label{trajopt_chance} \\
%& \bm{\mathcal{P} \bigg( \mathcal{G}_t \big(\mathcal{E}_t, y_t^1, \dots, y_t^M \big) < 0 \bigg) < \Delta}, & \nonumber \\
& \forall t = 1, 2, \dots, N, \;  \forall i = 1, 2, \dots, M, \nonumber  & 
\end{align}
where $ f $ is the nonlinear dynamic system model of robot, $ g^i $ is the prediction model of the $ i$-th dynamic obstacles, $ \mathcal{E}_t $ is the propagated FRS computed under bounded disturbance, $ \mathcal{F}_t $ is the collision free regions with regards to static structures, and $ \mathcal{U} $ is the feasible control set. The function $ \mathcal{G}_t(\cdot) $ encode the region that robot is in collision with dynamic obstacles. $ \mathcal{P} $ indicates the probability of an event. $ \Delta_t $ is the prescribed collision probability for step $ t $.
This work utilizes the Gaussian distribution and Forward Reachable Set (FRS) to address the uncertainties arising from dynamic obstacles and external disturbance, respectively.
By applying the chance constraint (\ref{trajopt_chance}), we ensure that the collision probability between dynamic obstacles and every state in FRS is below a user-specified threshold.

In this work, we make two key assumptions. Firstly, the noise of robot and obstacle models are all in the representation of Gaussian distribution. This is valid because of the central limit theorem as the robot system evolves over time. While the Gaussian distribution of obstacles' state encodes the belief of the prediction model.
Secondly, we assume that shapes of robots and dynamic obstacles are all in ellipsoids. Actually, for an arbitrary geometric shape, we can always find its minimum-volume enclosing ellipsoids, the \textit{L\"{o}wner-John ellipsoids}, by convex optimization \cite{rimon1997obstacle}.

\section{Tight Collision Probability}
\label{tcc}
\begin{figure}[h]
	\centering
	\includegraphics[scale=0.4]{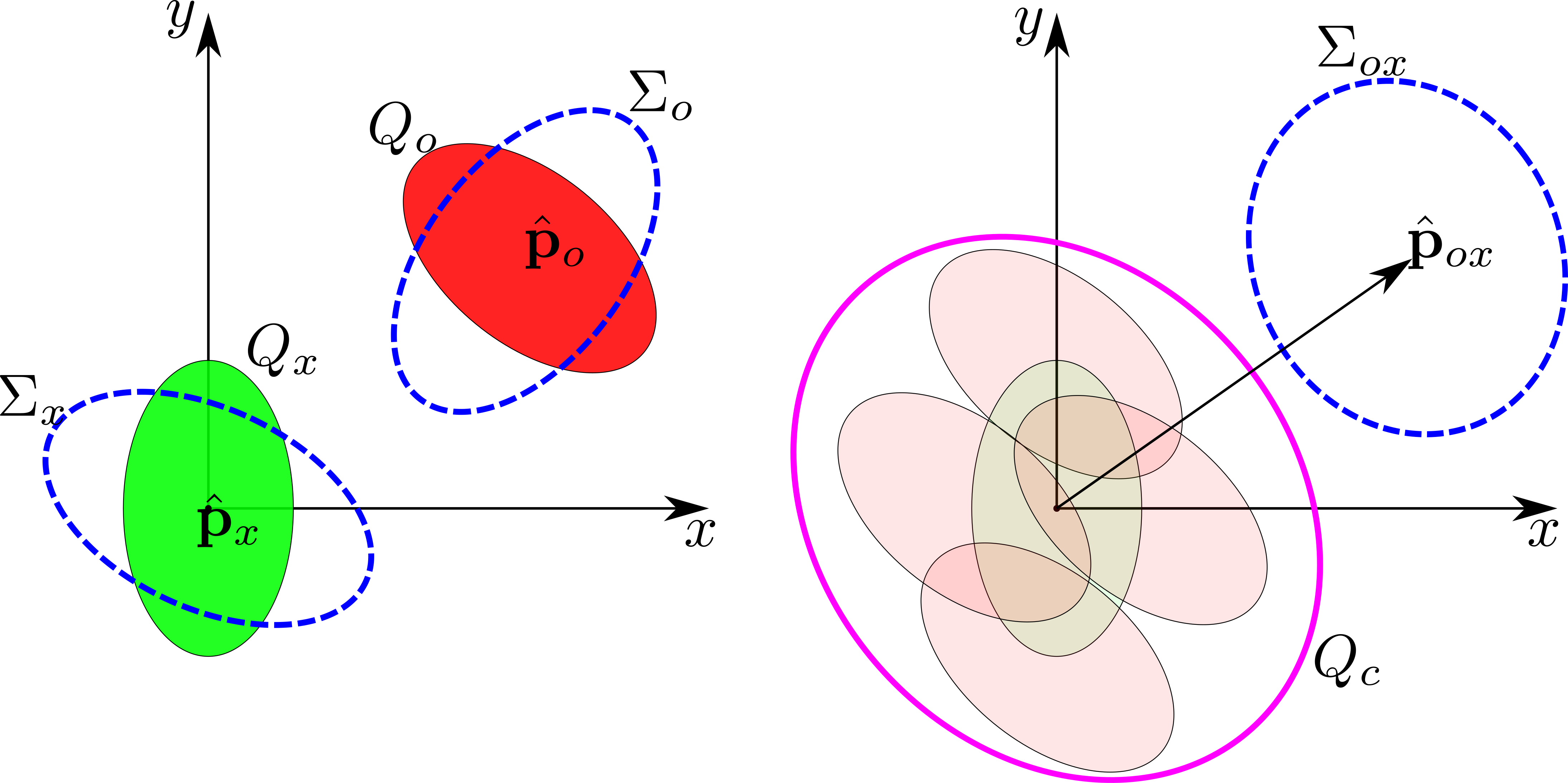}
	%	\hspace{1in}
	\caption{Illustration of the collision region upper bound $ Q_c $ (pink ellipse) in 2D. \textbf{Red ellipse}: obstacle shape. \textbf{Green ellipse}: robot shape. \textbf{Blue ellipse}: Gaussian distribution.}
	\label{coll_q_c}
	\vspace{-0.2cm}
\end{figure}
Suppose we have an ellipsoidal-shaped robot and obstacle with shape matrix of  $ Q_x, \: Q_o \in \mathbb{S}_{+}^3 $, respectively. The collision region in configuration space is thus the Minkowski sum of these two ellipsoids. As shown in \hyperref[coll_q_c]{Figure \ref{coll_q_c}}, we approximate the region with an optimal outer ellipsoid which is denoted by:
\begin{equation}
Q_c = Q_x \boxplus Q_o = (1 + \alpha)Q_x + (1 + \frac{1}{\alpha}) Q_o,
\end{equation}
where $ \alpha = \tr(Q_o)  / \tr(Q_x) $. 
Suppose the positions and uncertainty covariances of the robot and the obstacle are $ \mathbf{p}_x \sim \mathcal{N}(\hat{\mathbf{p}}_x, \Sigma_{x})$, $ \mathbf{p}_o \sim \mathcal{N}(\hat{\mathbf{p}}_o, \Sigma_{o}) $, with $ \mathbf{p}_x, \: \mathbf{p}_o \in \mathbb{R}^3 $. We can calculate the upper bound of the collision probability as:
\begin{equation}
 \mathcal{P}_c = \int_{\parallel \mathbf{p}_{o} - \mathbf{p}_{x} \parallel_{Q_c^{-1}} < 1 } (\mathbf{p}_{o} - \mathbf{p}_{x}) d(\mathbf{p}_{o} - \mathbf{p}_{x}) > \mathcal{P},
\label{ub_integral}
\end{equation}
where $ \mathbf{p}_o - \mathbf{p}_x = \mathbf{p}_{ox}  $ is the relative position between robot and obstacle, and $ \parallel \mathbf{p}_{ox} \parallel_{Q_c^{-1}} =  \mathbf{p}_{ox}^T Q_c^{-1} \mathbf{p}_{ox} < 1 $ is the outer ellipsoid $ Q_c $. The probability  (\ref{ub_integral}) is the integral of the relative position distribution $ \mathbf{p}_{ox}\sim \mathcal{N}(\hat{\mathbf{p}}_o - \hat{\mathbf{p}}_x,  \Sigma_{o} + \Sigma_{x}) $ over the ellipsoid $ Q_c $. 

In the following subsections, we present the approaches to evaluate $ \mathcal{P}_c $ and how to achieve the chance constraint (\ref{trajopt_chance}) in the form of 
\begin{equation}\label{tcc_e}
 \mathcal{P}_c \Big( \mathcal{G}_t \big(\mathcal{E}_t, y_t^1, \cdots y_t^M \big) < 0 \Big) < \Delta_t.
\end{equation}

%Many methods loosely estimate the probability in (\ref{trajopt_chance}), while this section detailed describe how we obtain a tight bound and how to iteratively reach the tight chance constraint. Benchmark comparisons are conducted at the end. 
\subsection{Collision Probability Estimation}
\label{coll_prob_est}

%Next , we show how to obtain the exact solution of (\ref{ub_integral}).

In the view of the random variable $ \mathbf{p}_{ox} $, the probability (\ref{ub_integral}) can be rewritten as 
\begin{equation}
\mathcal{P}_c = \mathcal{P}(\mathbf{p}_{ox}^T Q_c^{-1} \mathbf{p}_{ox} < 1) = \mathcal{P}(v < 1) = F_v(1), 
\label{ub_quadratic}
\end{equation}
where $ v = \mathbf{p}_{ox} Q_c^{-1} \mathbf{p}_{ox}^T $ is the quadratic form in random variable $ \mathbf{p}_{ox} $ and $ F_v $ is the cumulative distribution function (CDF) of $ v $. According to \cite{0Quadratic}, $ F_v $ has an exact expression in the form of series expansions, as stated by the following theorem.

\begin{theorem}\label{thm1}
\textit{
Suppose the quadratic form $ v = \mathbf{p}^T A \mathbf{p} $ with   $A=$ $A^T>0, \, \mathbf{p} \sim \mathcal{N}(\boldsymbol{\mu}, \Sigma), \Sigma>0$. With the affine transformation $ \mathbf{z} = \Sigma^{-\frac{1}{2}} \mathbf{p} - \Sigma^{-\frac{1}{2}} \boldsymbol{\mu} $ and the diagonalization $P^T \Sigma^{\frac{1}{2}} A \Sigma^{\frac{1}{2}} P=\operatorname{diag}\left(\lambda_1, \ldots, \lambda_n\right)$,  the quadratic form v can be expressed as:
$$
v = (\mathbf{w} + \mathbf{b})^T \diag (\lambda_1, \dots, \lambda_n) (\mathbf{w} + \mathbf{b}),
$$
where $ P $ is an orthogonal matrix ($P P^T=I$), $\mathbf{w}=P^T \mathbf{z}=\left(w_1, \dots, w_n\right)^T $ and $ \mathbf{b}=P^T \Sigma^{-\frac{1}{2}} \boldsymbol{\mu}=$ $\left(b_1, \ldots, b_n\right)^T$.
In this representation, the CDF of v is 
$$
F_{v}(q)=P(v \leq q)=\sum_{k=0}^{\infty}(-1)^k c_k \frac{q^{\frac{n}{2}+k}}{\Gamma\left(\frac{n}{2}+k+1\right)},
$$
where $\Gamma$ denotes the gamma function and 
$$
\begin{aligned}
c_0 & =\exp \left(-\frac{1}{2} \sum_{i=1}^n b_i^2\right) \prod_{i=1}^n\left(2 \lambda_i\right)^{-\frac{1}{2}}, \\
c_k & =\frac{1}{k} \sum_{i=0}^{k-1} d_{k-i} c_i, \\
d_k & =\frac{1}{2} \sum_{i=1}^n\left(1-k b_i^2\right)\left(2 \lambda_i\right)^{-k}.
\end{aligned}
$$
}
\end{theorem} 

In our case, the $ n = 3$ and $ q = 1 $. The convergent proof of this infinite series as well as the derivation of the truncation error upper bound, is presented in \cite{thomas2021integrated, kotz1967series}.

Additionally, the probability $ \mathcal{P}_c $ can be computed numerically via Gauss-Hermit quadrature. First of all, we apply the affine coordinate transform $ \mathbf{r} = R^T \mathbf{p}_{ox} $, where $ R $ is an orthogonal matrix that diagonalizes $ \Sigma_{ox} $, i.e. $ R^T\Sigma_{ox}R = \Lambda$. Thus, each dimension $ \mathbf{r}^i $ of $ \mathbf{r} $ is independent from each other. Under the new coordinates, $ \mathbf{p}_{ox} $ and $ Q_c $ become $ \mathbf{r} $ and $ Q_r $:
\begin{equation}\label{coor_trans}
\begin{aligned} &
\mathbf{r}   \sim \mathcal{N}(\hat{\mathbf{r}}, \Sigma_r) = \mathcal{N}(R^T \hat{\mathbf{p}}_{ox}, \Lambda), \\
& Q_r    = (RQ_c^{-1}R^T)^{-1}.
\end{aligned}
\end{equation}
Then the probability $ \mathcal{P}_c $ is rewritten as 
\begin{equation}\label{pc_trans}
\mathcal{P}_c = \int_{\mathbb{R}^3} I(\mathbf{r})p(\mathbf{r}) d\mathbf{r},
\end{equation}
where $ I(\cdot) $ is a function to indicate whether $ \mathbf{r} $ is in the collision region:
\begin{equation}\label{indicator_f}
I(\mathbf{r})= 
\begin{cases}
1, &  \mathbf{r}^TQ_r^{-1} \mathbf{r} < 1,  \\ 
0, & \text { otherwise. }
\end{cases}
\end{equation}
Since all the dimensions of $ \mathbf{r} $ are independent, we have : 
\begin{equation}\label{r_attr}
\begin{aligned}
p\left(\mathbf{r}\right) &= p(\mathbf{r}_1) p(\mathbf{r}_2) p(\mathbf{r}_3), \\
p ( \mathbf{r}_i ) & =\frac{1}{ \boldsymbol{\sigma}_i \sqrt{2 \pi}} \exp( -\frac{ (\mathbf{r}_i-\boldsymbol{\mu}_i )^2}{2 \boldsymbol{\sigma}_i^2} ),
\end{aligned}
\end{equation}
where $\boldsymbol{\mu} =  R^T \hat{\mathbf{p}}_{ox}$ and  $ \boldsymbol{\sigma}_i $ is the $ i-i $ entry of $ \Sigma_r $.
Applying (\ref{r_attr}) to (\ref{pc_trans}) yields:
\begin{equation}\label{pc_open}
\mathcal{P}_c = \int_{-\infty}^{\infty} p(\mathbf{r}_1) \int_{\mathbb{R}^{2}} p(\mathbf{r}_{2:3}) I(\mathbf{r}
) d \mathbf{r}_{2: 3} d \mathbf{r}_1 .
\end{equation}
Let $g(\mathbf{r}_1)=\int_{\mathbb{R}^{2}} p(\mathbf{r}_{2: 3}) I(\mathbf{r}) d \mathbf{r}_{2: 3}$, then:
\begin{equation}\label{pc_open2}
\begin{aligned}
\mathcal{P}_c & =\int_{-\infty}^{\infty} p(\mathbf{r}_1) g(\mathbf{r}_1) d \mathbf{r}_1 \\
& =\int_{-\infty}^{\infty} \frac{1}{\boldsymbol{\sigma}_1 \sqrt{2 \pi}} \exp (-\frac{(\mathbf{r}_1-\boldsymbol{\mu}_1)^2}{2 \boldsymbol{\sigma}_1^2}) g(\mathbf{r}_1) d \mathbf{r}_1.
\end{aligned}
\end{equation}

Gauss-Hermite quadrature approximates the value of integrals by calculating the weighted sum of the integrand function at a finite number of reference points, i.e.
\begin{equation}\label{gh_ori}
\int_{-\infty}^{\infty} e^{-z^2} h(z) d z \approx \sum_{j=1}^n w_j h\left(z_j\right),
\end{equation}
where $n$ is the number of sampled points, $z_j \, (j = 1, 2, \ldots, n)$ are the roots of the Hermite polynomial $H_n(z)$ and the associated weights $w_j$ are given by:
\begin{equation}\label{gh_w}
w_j=\frac{2^{n-1} n ! \sqrt{\pi}}{n^2\left[H_{n-1}\left(z_j\right)\right]^2} .
\end{equation}
To match (\ref{pc_open2}) to (\ref{gh_ori}), we take the following variable change:
\begin{equation}\label{var_trans}
\mathbf{z}_1=\frac{\mathbf{r}_1-\boldsymbol{\mu}_1}{\sqrt{2} \boldsymbol{\sigma}_1} \Leftrightarrow \mathbf{r}_1=\sqrt{2} \boldsymbol{\sigma}_1 \mathbf{z}_1+\boldsymbol{\mu}_1,
\end{equation}
such that the Equation (\ref{pc_open2}) becomes
$$
\mathcal{P}_c=\int_{-\infty}^{\infty} \frac{1}{\sqrt{\pi}} e^{-\mathbf{z}_1^2} g(\sqrt{2} \boldsymbol{\sigma}_1 \mathbf{z}_1+\boldsymbol{\mu}_1) d \mathbf{z}_1.
$$
Hence, the value of $\mathcal{P}_c$ can then be approximated through Gauss-Hermite quadrature rule:
\begin{equation}\label{pc_gh}
\mathcal{P}_c \approx \frac{1}{\sqrt{\pi}} \sum_{j=1}^{n_1} w_{1, j} g\left(\sqrt{2} \boldsymbol{\sigma}_1 z_{1, j}+\boldsymbol{\mu}_1\right).
\end{equation}
%where $y_{1, j}\left(j=1, \ldots, n_1\right)$ are the Hermite polynomial roots for integrating the $x_t^1$ component, $w_{1, j}$ are the associated weights, and $n_1$ is the number of sampled points.
By iteratively applying this procedure from $\mathbf{r}_1$ to $\mathbf{r}_3$, we can obtain a numerical integral of $ \mathcal{P}_c $:
\begin{equation}\label{pc_app}
\begin{aligned}
\mathcal{P}_c \approx & \pi^{-\frac{3}{2}} \sum_{j_1=1}^{n_1} \sum_{j_2=1}^{n_2} \sum_{j_3=1}^{n_3} \left(\prod_{i=1}^d w_{i, j_i}\right) I ( [ \sqrt{2} \boldsymbol{\sigma}_1 z_{1, j_1}  \\
& +\boldsymbol{\mu}_1, 
\sqrt{2} \boldsymbol{\sigma}_2 z_{2, j_2} +\boldsymbol{\mu}_2, \sqrt{2} \boldsymbol{\sigma}_3 z_{3, j_3}+\boldsymbol{\mu}_3  ]^T ).
\end{aligned}
\end{equation}

\subsection{Iterative Trajectory Optimization}
\label{ite_traj_opt}

\begin{figure}[h]
	\centering
	\vspace{0.1cm}
	\includegraphics[scale=0.6]{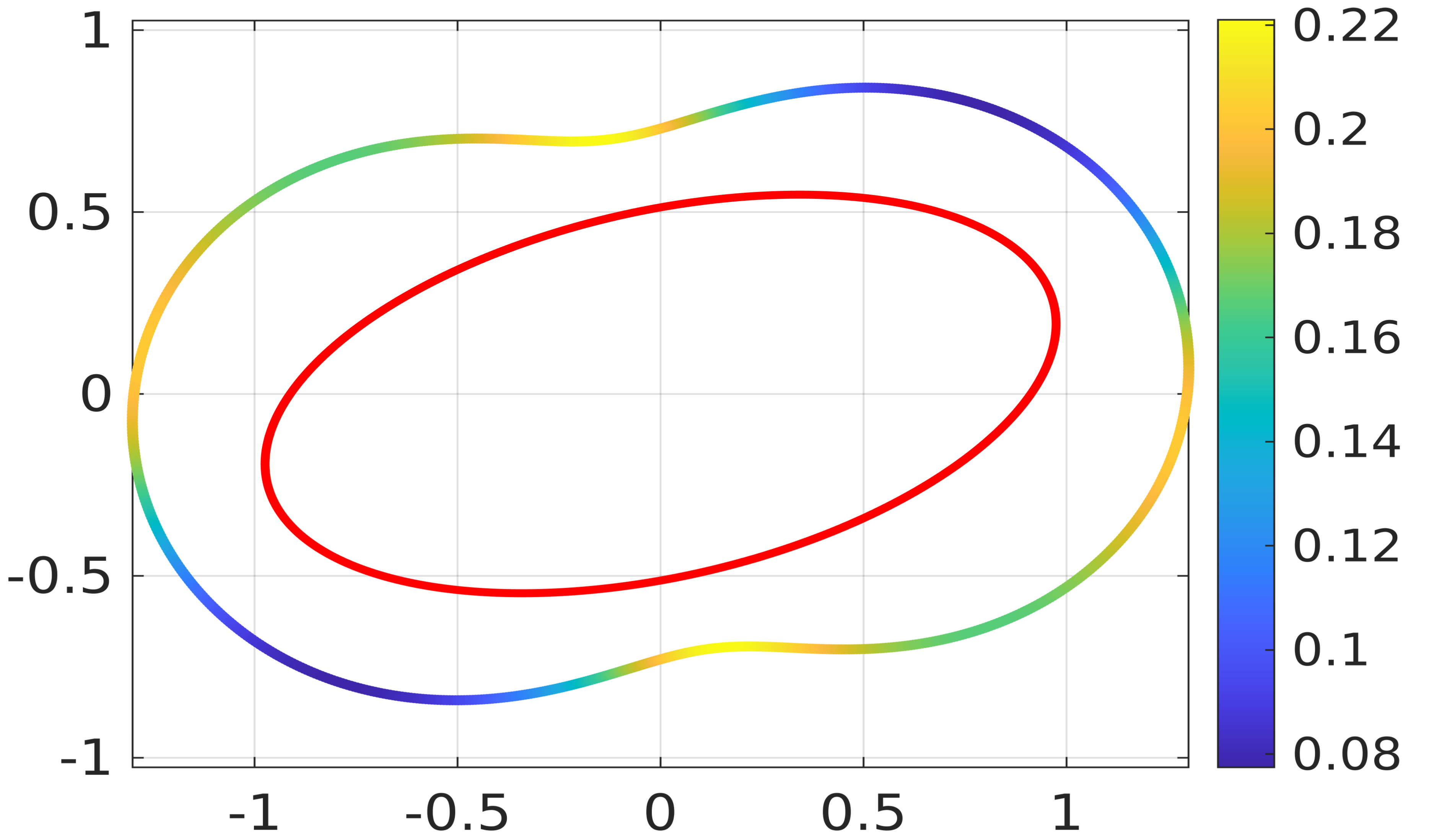}
	%	\hspace{1in}
	\caption{0.3-level set for the linearized collision probability in 2D. The red ellipse indicates the collision region, and the color on level set encodes the ground truth probability value.}
	\label{l_level_set}
\end{figure}
The linearized collision probability $ \mathcal{P}_l $ proposed in \cite{zhu2019chance} has the ability to convert the chance constraint $ \mathcal{P}_l < \Delta $ into a deterministic one, thus enabling efficient trajectory optimization. We summarize the method in \hyperref[conlu1]{Conclusion \ref{conlu1}}.
\begin{conclusion}\label{conlu1}
	\textit{ Let the collision region in configuration space be the ellipsoid $ \mathbf{p}^T Q^{-1} \mathbf{p} < 1 $ with $ \mathbf{p} \in \mathbb{R}^3, Q \in \mathbb{S}_{+}^3 $. Denote the robot-obstacle relative position distribution as $ \mathbf{p} \sim  \mathcal{N}(\hat{\mathbf{p}},  \Sigma) $. The linearized collision probability $ \mathcal{P}_l $ is given by
		$$
		\mathcal{P} < \mathcal{P}_l = \frac{1}{2} + \frac{1}{2}\erf \left( \frac{1-\mathbf{a}^TQ^{- \frac{1}{2}}\hat{\mathbf{p}}   }{\sqrt{ 2\mathbf{a}^T Q^{- \frac{1}{2}} \Sigma Q^{- \frac{1}{2}} \mathbf{a}  }} \right), 
		$$
	where $ \mathbf{a} = Q^{- \frac{1}{2}} \hat{\mathbf{p}} / \|  Q^{- \frac{1}{2}} \hat{\mathbf{p}} \|  $. The chance constraint $ \mathcal{P}_l < \Delta $ can then be transformed to the a deterministic one:
	$$
	\mathbf{a}^TQ^{- \frac{1}{2}}\hat{\mathbf{p}}  - 1 > \erf^{-1}(1-2\Delta) \sqrt{ 2\mathbf{a}^T Q^{- \frac{1}{2}} \Sigma Q^{- \frac{1}{2}} \mathbf{a}  }.
	$$
	}
where $ \erf(\cdot) $ is the error function: $ \erf(x) = \frac{2}{\sqrt{\pi}} \int_0^x e^{-t^2} dt $ .
\end{conclusion} 

 However, as illustrated in \hyperref[l_level_set]{Figure \ref{l_level_set}}, this method over-estimates the probability and gives rise to a loose chance constraint. And our exact solution of $ \mathcal{P}_c $ equals the ground truth value, while the true estimation of $ \mathcal{P}_l $ ranges from $ 0.0774 $ to $ 0.2211 $ for the $ 0.3 $-level set. By combining with the linearized collision probability, we propose an iterative trajectory optimization method to achieve the tight constraint (\ref{tcc_e}).
 
\begin{algorithm}[tp]
	\footnotesize
	\DontPrintSemicolon
	\SetKwFunction{FMain}{\textbf{IterTrajOpt}}
	\SetKwProg{Pn}{Function}{:}{\KwRet}
	\KwIn{Total allowed risk $ \Delta $, the optimized steps $ n $}
	\KwResult{the optimized trajectory $ \Gamma $}
	\Pn{\FMain{}}{
		$ \Delta_l \leftarrow 0, \; \delta_l \leftarrow 0$  \;
		$ \Delta_h \leftarrow n, \; \delta_h \leftarrow 0$  \;
		$ \Gamma_0 \leftarrow $ trajectory optimization without chance constraint\;
		$ \delta_0 \leftarrow$ get trajectory risk  of $ \Gamma_0 $ \;
		$ \delta_h \leftarrow \delta_0 $ \;
		\lIf{$ \delta_0 \le \Delta $ }{\Return{$ \Gamma_0 $}}
		\;
		$ k \leftarrow 0 $ \;
		\While{$ |\delta_k - \Delta| >$ precision}{
			$ k \leftarrow k+1 $ \;
			$ \Delta_k \leftarrow  \Delta_l + (\Delta_h - \Delta_l)/(\delta_h - \delta_l) (\delta_{k-1} - \delta_l) $ \;
			$ \Gamma_k \leftarrow $ trajectory optimization with $ \mathcal{P}_l <  \Delta_k / n $\;
			$ \delta_k \leftarrow$ get trajectory risk  of $ \Gamma_k $ \;
			\lIf{$ \delta_k < \Delta $}{$ \Delta_l = \Delta_k, \; \delta_l = \delta_k $}
			\lElse{$ \Delta_h = \Delta_k, \; \delta_h = \delta_k $}
		}
		
		\Return{$ \Gamma_k $}
	}
	\caption{Iterative Trajectory Optimization}
	\label{alg1}
\end{algorithm}
In particular, given a trajectory $\Gamma = (\mathbf{x}_1, \dots, \mathbf{x}_n)  $, we take uniform risk allocation $ \Delta_t = \Delta / n $ for every step. The collision probability for each step is taken as the maximum one among all obstacles.
As mentioned in \cite{frey2020collision}, summing the collision probabilities at discrete time steps will double count the probability. Thus the risk here (i.e. the sum of all probabilities in different steps) is not necessarily to be lower than one.  Notice that trajectory optimization under linearized probability $ \mathcal{P}_l $
tightens the original problem, while trajectory optimization without chance constraint is a relaxation. Hence our goal is to achieve the tight chance constraint (\ref{tcc_e}) between them. 
As shown in \hyperref[alg1]{Algorithm \ref{alg1}}, trajectory optimization with $ \mathcal{P}_l < \Delta/n $ is solved iteratively. After each solving step, the exact solution of $ \mathcal{P}_c $ is employed to evaluate the risk of optimized trajectory and provide a guess of the next suitable value of $ \mathcal{P}_l $. Observing that $ \mathcal{P}_l $ has positive correlation with tight probability $ \mathcal{P}_c $ in general, we guess the next value of 
$ \mathcal{P}_l $ via proportional interpolation. It has been verified in our experiments that this heuristic can help us quickly reach the tight chance constraint. To reduce the computation time,  the previously obtained solution can serve as an initial guess for the next iteration of the problem.

\subsection{Benchmark Comparisons}
\begin{figure}[ht]
	\centering
	\includegraphics[scale=0.69]{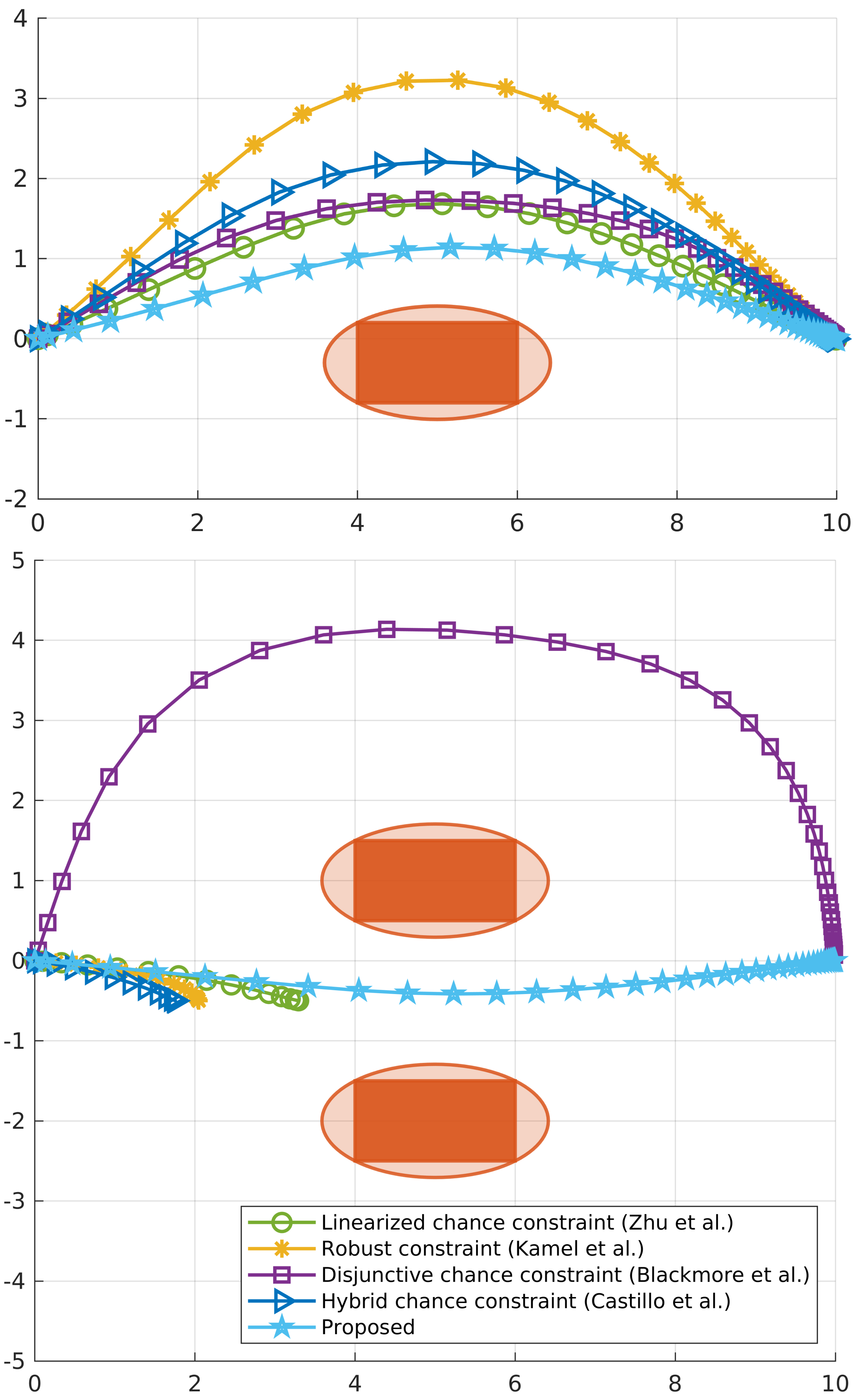}
	%	\hspace{1in}
	\caption{Benchmark comparison of our tight chance constraint against the linearized chance constraint \cite{zhu2019chance}, robust constraint \cite{kamel2017robust}, disjunctive chance constraint \cite{blackmore2011chance} and hybrid chance constraint \cite{castillo2020real}. }
	\label{opt_bench}
\end{figure}
%We statistically compare our collision probability estimation methods with others. Let the mean position of the first ellipsoid be at the origin, while the mean position of the second ellipsoid is uniformly generated within a $ 4 \times 4 \times 4 $ cube centered at the origin. The covariance matrices are also randomly generated with $ \sigma^2 \in [0.01, 2] $ for each independent dimension. The semi-axis length of the two ellipsoid are range in $ [0.2, 2]$. The first ellipsoid is axis aligned and the orientation of the second one is random generated. We benchmark our methods in this setting for ten thousand cases and obtain results in \hyperref[prob_bench_res]{Table \ref{prob_bench_res}}. It shows that our methods achieve the best results in the limited time. Although fast Monte Carlo has the most accurate result, it takes a long time for sampling. Compared to the exact solution, our approximation method with $ n =10 $ can give a good result in little time, but pay more effort to improve the accuracy. Therefore, the method is adopted to check motion primitives to provide a reference trajectory for optimization.

We statistically compare our collision probability estimation methods with others\footnote{The relative performance differs from the result presented in \cite{castillo2020real}; refer to \href{https://github.com/Acmece/IROS2023_Com}{https://github.com/Acmece/IROS2023\_Com} for an explanation.} in a setting where the mean position of the first ellipsoid is at the origin, the mean position of the second ellipsoid is uniformly generated within a $4 \times 4 \times 4$ cube centered at the origin, and the covariance matrices are randomly generated with $ \sigma^2 \in [0.01, 2]$ for each independent dimension. The semi-axis length of the two ellipsoids ranges in $[0.2, 2]$; the first ellipsoid is axis aligned and the orientation of the second one is randomly generated. We evaluate and benchmark our methods in this setting through ten thousand cases and the resulting performance is presented in  \hyperref[prob_bench_res]{Table \ref{prob_bench_res}}, showing that our methods achieve the best result in the limited time. Although the fast Monte Carlo method has the most accurate result, it requires a significant amount of time for sampling.  On the other hand, our approximation method with a low number of quadrature points ($n=10$) can provide a good result in a short amount of time, but further efforts are required to enhance its accuracy to match that of the exact method. Thus, the method is adopted to check motion primitives to provide a reference trajectory for optimization.
\begin{table}
	\centering
	\fontsize{6}{11} \selectfont
	\vspace{0.2cm}
	\caption{Benchmark statistics on collision probability Estimation} 
	\begin{tabular}{c|c c}
		\toprule
		\toprule
		Methods & Collision Probability Error & Computation Time (s) \cr
		\hline
		Fast Monte Carlo \cite{lambert2008fast} & -0.0020 $  \pm $  0.0024 & 4.3425 $ \pm $ 0.9781 \cr
		Max Point Approximation \cite{park2018fast} & 0.5483 $ \pm $ 0.3957 & 0.0883 $ \pm $ 0.0361 \cr
		Center Point Approximation \cite{du2011probabilistic} & 0.3390 $ \pm $ 0.2390 & 0.0004 $ \pm $ 0.0002  \cr
		Linearized Probability \cite{zhu2019chance} & 0.4314 $ \pm $ 0.1630 & 0.0016 $ \pm $ 0.0008 \cr
		Rectangular Bounding Box \cite{hardy2013contingency} & 0.3224 $ \pm $ 0.1547 & 0.0035 $ \pm $ 0.0011 \cr
		Sphere Approximation \cite{thomas2021integrated} & 0.4928 $ \pm $ 0.1885 & 0.0066 $\pm $ 0.0019  \cr
		Markov's Inequality \cite{thomas2021exact} &  0.5479 $ \pm $0.7966  & 0.0013 $ \pm $ 0.0003 \cr
		Ours-Approximate ($ n = 10 $) & 0.1799 $ \pm $ 0.1925 & 0.0018 $ \pm $ 0.0010 \cr
		Ours-Approximate ($ n = 200 $) & 0.1372 $ \pm $ 0.1432  & 0.0201 $ \pm $ 0.0065\cr
		Ours - Exact  & 0.1257 $ \pm 0.1364 $ & 0.0063 $ \pm $ 0.0032	   \cr
		\bottomrule
		\bottomrule		   
	\end{tabular}
	\label{prob_bench_res}
\end{table}	

%\begin{table}
%	\centering
%	\fontsize{6}{11} \selectfont
%	\vspace{0.2cm}
%	\caption{Benchmark statistics on collision probability Estimation} 
%\begin{tabular}{c|c c}
%	\toprule
%	\toprule
%	Methods & Collision Probability Error & Computation Time (s) \cr
%	\hline
%	Fast Monte Carlo [1] & -0.0020 $  \pm $  0.0024 & 4.3425 $ \pm $ 0.9781 \cr
%	Max Point Approximation [2] & 0.5483 $ \pm $ 0.3957 & 0.0883 $ \pm $ 0.0361 \cr
%	Center Point Approximation [3] & 0.3390 $ \pm $ 0.2390 & 0.0004 $ \pm $ 0.0002  \cr
%	Linearized Probability [4] & 0.4314 $ \pm $ 0.1630 & 0.0016 $ \pm $ 0.0008 \cr
%	Rectangular Bounding Box [5] & 0.3224 $ \pm $ 0.1547 & 0.0035 $ \pm $ 0.0011 \cr
%	Sphere Approximation [6] & 0.4928 $ \pm $ 0.1885 & 0.0066 $\pm $ 0.0019  \cr
%	Markov's Inequality [7] &  0.5479 $ \pm $0.7966  & 0.0013 $ \pm $ 0.0003 \cr
%	Ours-Approximate ($ n = 10 $) & 0.1799 $ \pm $ 0.1925 & 0.0018 $ \pm $ 0.0010 \cr
%	Ours-Approximate ($ n = 200 $) & 0.1372 $ \pm $ 0.1432  & 0.0201 $ \pm $ 0.0065\cr
%	Ours - Exact  & 0.1257 $ \pm 0.1364 $ & 0.0063 $ \pm $ 0.0032	   \cr
%	\bottomrule
%	\bottomrule		   
%\end{tabular}
%\label{prob_bench_res}
%\end{table}	

\begin{table}
	\centering
	\fontsize{8}{11} \selectfont
	\caption{Benchmark statistics on one horizon planning} 
	\begin{tabular}{cc|ccccc}
		\toprule
		\toprule
		\multicolumn{2}{c|}{Scenes}  & \textbf{Proposed} & \textbf{Zhu} & \textbf{Kamel} & \textbf{Blackmore} & \textbf{Castillo} \cr
		
		\hline
		
		\multirow{3}{*}{\uppercase\expandafter{\romannumeral1}} & Obj & \textbf{1.00}  & 1.03 & 1.22 & 1.05 & 1.09  \cr
		& Time & 1.00  & 0.83 & 0.52 & 119.26 & \textbf{0.46}  \cr
		& Risk & \textbf{0.4001}  & 0.0530 & 0 & 0.0314 & 0.0037  \cr
		
		\hline
		\multirow{3}{*}{\uppercase\expandafter{\romannumeral2}} & Obj & \textbf{1.00}  & 2.03 & 2.41 & 1.69 & 2.65  \cr
		& Time & 1.00  & 0.31 & 0.27 & 6246.49 & \textbf{0.22}  \cr
		& Risk & \textbf{0.0107}  & 0.0038 & 0 & 0 & 0  \cr
		
		\bottomrule
		\bottomrule		   
	\end{tabular}
	\label{one_plan_bench_res}
	\vspace{-0.5cm}
\end{table}

The iterative trajectory optimization algorithm is tested in two one-horizon planning scenes. A robot of single mass point starts from the origin and is expected to reach the goal at $ [10, 0] $. All the obstacles are within a bounding box of size $ [1, 0.5] $. In the first scene, there is only one obstacle with mean position $ \mathbf{p} = [5, -0.3]^{\mathrm{T}} $ and covariance $ \Sigma = \diag(0.5, 0.3) $. The second scene has two obstacles and the mean positions and covariances are $ \mathbf{p}_1 = [5, -2]^{\mathrm{T}} $,  $ \mathbf{p}_2 = [5, 1]^{\mathrm{T}} $ , $ \Sigma_1 = \diag(0.4, 0.07) $, $ \Sigma_2 = \diag(0.4, 0.07) $, respectively. We set a planning horizon of $ 8 $ seconds with $ N = 40 $ steps. The prescribed collision risk is $ 0.4 $ for scene \uppercase\expandafter{\romannumeral1} and $ 0.01 $ for scene \uppercase\expandafter{\romannumeral2}. As shown in \hyperref[opt_bench]{Figure \ref{opt_bench}} and \hyperref[one_plan_bench_res]{Table \ref{one_plan_bench_res}}, our iterative trajectory optimization method achieves the least conservative results compared to all the other methods in scene \uppercase\expandafter{\romannumeral1}.  In scene \uppercase\expandafter{\romannumeral2}, only our method can go through the narrow gap and all the other methods fall into the local minima except \cite{blackmore2011chance}. Although the method \cite{blackmore2011chance} can find the global optimal solution, but its use of the linearized collision probability leads to an overly conservative solution. All these methods are implemented in Matlab using CasADi framework \cite{Andersson2018}.
 Our method iterates three times for scene \uppercase\expandafter{\romannumeral1} while spending a comparable time to \cite{zhu2019chance}. In scene \uppercase\expandafter{\romannumeral2}, our method iterates
 two times and the methods \cite{kamel2017robust,blackmore2011chance,castillo2020real} early exit because of reaching local minima.

%\begin{table}[h!]
%	\begin{center}
%		\caption{One Horizon Benchmark Results.}
%		\label{tab:table1}
%		\begin{tabular}{c|ccccc} 
%			& \textbf{Proposed} & \textbf{Zhu} & \textbf{Kamel} & \textbf{Blackmore} & \textbf{Manuel}  \\
%			\hline
%			Objective & 1 & 1.03 & 1.22 & 1.05 & 1.09  \\
%			CPU time & 1 & 0.70 & 0.484 & 119.260 & 0.392 \\
%			Risk & 0.4001 & 0.0530 & 0 & 0.0314  & 0.0037
%		\end{tabular}
%	\end{center}
%\end{table}
%
%\begin{table}[h!]
%	\begin{center}
%		\caption{One Horizon Benchmark Results.}
%		\label{tab:table1}
%		\begin{tabular}{c|ccccc} 
%			& \textbf{Proposed} & \textbf{Zhu} & \textbf{Kamel} & \textbf{Blackmore} & \textbf{Manuel}  \\
%			\hline
%			Objective & 1 & 2.0 & 2.4 & 1.6 & 2.6  \\
%			CPU time & 1 & 0.31 & 0.27 & 6246.49 & 0.22 \\
%			Risk & 0.0107 & 0.0038 & 0 & 0  & 0
%		\end{tabular}
%	\end{center}
%\end{table}

\section{UAV Motion Planning}
In this section, we implement the motion planning framework (\ref{trajopt}) in the Model Predictive Control (MPC) fashion for a quadrotor and conduct experiments in both simulation and real world.
\label{ump}
\subsection{Robot model}
We consider the dynamics model with accurate control of Euler angles, which assumes that its rate can accurately track the desired command. The system state is $\mathbf{x}=[\mathbf{p}, \mathbf{v}, \phi, \theta, \psi]^T \in$ $\mathcal{X} \subset \mathbb{R}^{n_x}$, where $\mathbf{p}=\left[p_x, p_y, p_z\right]^T, \mathbf{v}=\left[v_x, v_y, v_z\right]^T \in \mathbb{R}^3$ denotes the position and velocity of the quadrotor. $\phi, \theta, \psi \in \mathbb{R}$ are the roll, pitch and yaw angles. The control input is $\mathbf{u}=\left[\dot{\phi}_c, \dot{\theta}_c, \dot{\psi}_c, f_c\right]^T \in \mathcal{U} \subset \mathbb{R}^{n_u}$ in which $\dot{\phi}_c, \dot{\theta}_c, \dot{\psi}_c \in \mathbb{R}$ are the command rates of the Euler angles, $f_c \in \mathbb{R}$ is the total thrust command of the quadrotor in the body frame. We model the quadrotor dynamics as follows:
\begin{equation}\label{uav_model}
\begin{gathered}
\dot{\mathbf{p}}  =\mathbf{v} \\
\dot{\phi}  =\dot{\phi}_c,\dot{\theta}  =\dot{\theta}_c, \\
\dot{\mathbf{v}}  =\frac{1}{m}
\left(R 
\left[
\begin{array}{c}
0 \\
0 \\
f_c
\end{array}
\right] % +\mathbf{F}_{e x t}
\right) 
-\left[
\begin{array}{c}
0 \\
0 \\
g
\end{array}
\right],
\end{gathered}
\end{equation}
where $R \in \mathbb{R}^{3 \times 3}$ is the rotation matrix parameterized by the Euler angles, $g \in \mathbb{R}$ is the magnitude of gravitational acceleration, and $m \in \mathbb{R}$ is the mass of the quadrotor. 
The discrete
dynamics in (\ref{trajopt_robot}) is obtained through 4-th order Runge-Kutta integration of (\ref{uav_model}).

%Instead of predicting the external forces or modeling the field of special force like winds, we generally obtain a realtime force estimation and then adjust the planning strategy when the force surpasses its bound. The disturbance of external forces can be expressed as a nominal value with an additive bounded noise, which makes it possible to design an NMPC that satisfies the constraints under bounded disturbance. 
%The nominal force $\boldsymbol{b}_{e x t}$ can be treated as a constant value calculated in [9] for a sufficiently short duration. The external force is defined as $\boldsymbol{F}_{\text {ext }}:=\boldsymbol{b}_{\text {ext }}+\boldsymbol{w}_{\text {ext }}$, where the bounded noise $\boldsymbol{w}_{\text {ext }} \in \mathbb{W}=\left\{\boldsymbol{w} \in \mathbb{R}^{n_w}:\|\boldsymbol{w}\|_{\infty} \leq w_m\right\}, w_m$ is the maximum bound.

\subsection{Dynamic Obstacle Model}
%As reported in \cite{scholler2020constant}, the constant velocity model outperforms most of other prediction models in many aspects. Therefore, we model the dynamic obstacles as:

The constant velocity model is used for dynamic obstacle prediction because it outperforms most other models in many aspects \cite{scholler2020constant}. Thus, defining the obstacle state as $\mathbf{y}_t^i=\left[\mathbf{q}_t^i, \mathbf{v}_t^i, \phi_t^i, \theta_t^i,  \psi_t^i, \dot{\phi}_t^i, \dot{\theta}_t^i, \dot{\psi}_t^i \right]$, the constant velocity model is given by:
\begin{equation}\label{cvm}
\begin{aligned}
\dot{\mathbf{q}}^i &=R\left(\phi^i, \theta^i, \psi^i\right) \mathbf{v}^i, \\
\dot{\mathbf{v}}^i &= \mathbf{0}, \, \ddot{\phi}^i = \ddot{\theta}^i = \ddot{\psi}^i=0,
\end{aligned}
\end{equation}   
where $ \mathbf{q}^i $ is the position, $\phi^i, \theta^i, \psi^i$ are the roll pitch and yaw angles respectively, and $\mathbf{v}^i$ is the velocity in the body frame.
The discrete dynamics $ g^i(\mathbf{y}^i_t)  $ in (\ref{trajopt_obs}) is obtained by Euler integration of (\ref{cvm}).

%where $v^i$ and $\psi^i$ are the linear velocity in the body frame and yaw angle of the $i$-th obstacle respectively. Thus, obstacle states are defined as $y_t^i=\left[\begin{array}{llll}q_t^i & v_t^i & \psi_t^i & \dot{\psi}_t^i\end{array}\right]$ where the nominal discrete dynamics $g^i\left(y_t^i\right)$ are determined through Euler integration of (20). 

%Similarly, the nominal state $\hat{y}_t^i$ and its covariance matrix $\Sigma_t^{y^i}$ are approximated with a first-order Taylor expansion
%$$
%\begin{aligned}
%& \hat{y}_{t+1}^i=g^i\left(\hat{y}_t^i\right) \\
%& \Sigma_{t+1}^{y^i}=\nabla g^i\left(\hat{y}_t^i\right) \Sigma_t^{y^i}\left(\nabla g^i\left(\hat{y}_t^i\right)\right)^T+V_t^i
%\end{aligned}
%$$

\subsection{Uncertainty Propagation}
To evaluate the chance constraint (\ref{trajopt_chance}), the uncertainty covariances of the states in (\ref{trajopt_robot}), (\ref{trajopt_obs}) and the FRS in (\ref{trajopt_frs}) need to be propagated over time. 
For the uncertainty of the robot and obstacle states,
since in our case the
planning horizon is short, we
propagate the covariances using an EKF-like update to achieve real time performance:
\begin{equation}\label{uncer_propa}
\begin{gathered}
\Sigma_{x,t+1} = F_{x,t} \Sigma_{x,t} (F_{x,t})^T + W_t, \\
\Sigma^{i}_{y,t+1} = F^{i}_{y,t} \Sigma^{i}_{y,t} (F^{i}_{y,t})^T + V_t,\\
\end{gathered}
\end{equation}
where $F_{x,t} = \left. \frac{\partial f}{\partial \mathbf{x}_t} \right|_{\hat{\mathbf{x}}_t, \mathbf{u}_t}$ and $F^{i}_{y,t} = \left. \frac{\partial g}{\partial \mathbf{y}_t^i} \right|_{\hat{\mathbf{y}}_t^i}$.
For the uncertainty results from external disturbances, we use \cite{seo2019robust} to calculate the FRS in a representation of ellipsoid $ Q_d $.

%E. Approximate Uncertainty Propagation
%Evaluating the chance constrainsts 6c and requires calculating the uncertainty covariance at each time step, i.e. uncertainty propagation. There are many methods to perform uncertainty propagation for nonlinear systems, for example the unscented transformation [19] and polynomial chaos expansions [20]. The readers can refer to [21] to get a comprehensive review. However, these methods are mostly computationally intensive and only outperform linearization methods when the propagation time is very long. In our case where the planning horizon is short, to achieve real time performance, we propagate uncertainties using a EKF-type update, i.e. $\Gamma_i^{k+1}=$ $F_i^k \Gamma_i^k F_i^{k^T}+Q_i^k$, where $\Gamma_i^k$ is the state uncertainty covariance at time $k, Q_i^k$ is the process noise and $F_i^k=\left.\frac{\partial \mathbf{f}_i}{\partial \mathbf{x}_i}\right|_{\hat{\mathbf{x}}_i^{k-1}, \mathbf{u}_i^k}$ is the state transition matrix of the robot. We further denote by $\Sigma_i^k$ the $3 \times 3$ covariance matrix of the position $\mathbf{p}_i^k$, extracted from $\Gamma_i^k$

\subsection{Cost Function}
We define the cost function as: 
\begin{equation}\label{cost}
\begin{aligned}
J = & \left\| \mathbf{p}_N-\mathbf{p}_N^{ref} \right \|_{l_p^N}  + \sum_{t=1}^{N-1} \left\| \mathbf{p}_t-\mathbf{p}_t^{ref} \right\|_{l_p}   \\
& 
+ \sum_{t=1}^{N}  \left\| \mathbf{u}_{t-1} \right\|_{l_u}  + 
\sum_{t=1}^{N-1} 
+  \left\| \mathbf{u}_t - \mathbf{u}_{t-1} \right\|_{l_\Delta u}
\end{aligned}
\end{equation}
where $ l_p^N $, $ l_p $, $ l_u $, $ l_{\Delta u} $ are norm-related weighting matrixes, and $ \mathbf{p}^{ref} $ is the position obtained from the reference trajectory. The reference trajectory is generated using a kinodynamic hybrid-state A* algorithm proposed in \cite{zhou2019robust}, which samples in the control space to generate motion primitives. We assess the risk of these primitives using the approximation method proposed in Section \ref{coll_prob_est}. The cost function (\ref{cost}) is designed to penalize the trajectory for tracking performance, energy efficiency, and smoothness.

\subsection{Collision Free Constraint}
For the static structure-related collision-free constraint in (\ref{trajopt_feas}),
we construct the flight corridor, a list of connected convex polyhedrons, using the method presented in  \cite{liu2017planning}. The flight corridor indicates the collision free region and we always constrain the FRS-augmented robot shape $ Q_a = Q_x \boxplus Q_d $  within it to ensure safety under bounded disturbance. Specifically, we denote $Q_a$ at time $t$ as $ Q_{a,t} $ and then assign a polyhedron $ \{A\mathbf{p} \leq \mathbf{b} \}$ to constrain it, with $ A \in \mathbb{R}^{n\times3}, \mathbf{b}\in \mathbb{R}^n$. Then, the point on the ellipsoid $ Q_{a,t} $, with the minimum signed distance to the $ i $-th plane of the polyhedron,  is 
\begin{equation}\label{min_point}
\setlength{\abovedisplayskip}{3pt}
\setlength{\belowdisplayskip}{3pt}
\mathbf{z}_t^i = \mathbf{p}_t + 
\frac{Q_{a,t} A_i^{\mathrm{T}}}{\left\| Q^{1/2}_{a,t} A_i^{\mathrm{T}} \right\|},
\end{equation}
where $ A_i $ is the $i$-th row of $ A $ and $ \mathbf{p}_t $ is the center of the ellipsoid $ \mathbf{Q}_{a,t} $.
%\frac{Q^{1/2}_{a,t} A_i}{\left\| Q^{1/2}_{a,t} A_i \right\|}
We achieve the collision-free constraint in (\ref{trajopt_feas}) by enforcing all such points $ \mathbf{z}_t^i $ to locate within their corresponding polyhedrons:
\begin{equation}\label{col_free_constraint}
A\mathbf{z}^{i}_t \leq \mathbf{b} \Leftrightarrow 
A\mathbf{p}_t + \left\|  Q^{1/2}_{a,t}A \right\| < \mathbf{b}.
\end{equation}

\subsection{Chance Constraint}
The chance constraint, as defined in equation (\ref{trajopt_chance}), is designed to maintain the probability of collision below a user-specified threshold over a forthcoming time horizon.
We use our proposed method in Section \ref{ite_traj_opt} to iteratively reach the tight chance constraint (\ref{tcc_e}). 
Since the robot shape is augmented with the FRS $ Q_d $, the collision region now becomes $ Q_c = Q_x \boxplus Q_o \boxplus Q_d $.
For real-time planning, we limit the iterative optimization time to a prescribed value. Additionally, we set $ \Delta_t = \Delta $ in \hyperref[alg1]{Algorithm \ref{alg1}} and solve it in the first iteration, so that the worst case result is that we reach the optimized trajectory with linearized collision probability.

\begin{figure}
	\centering
	\includegraphics[scale=0.14]{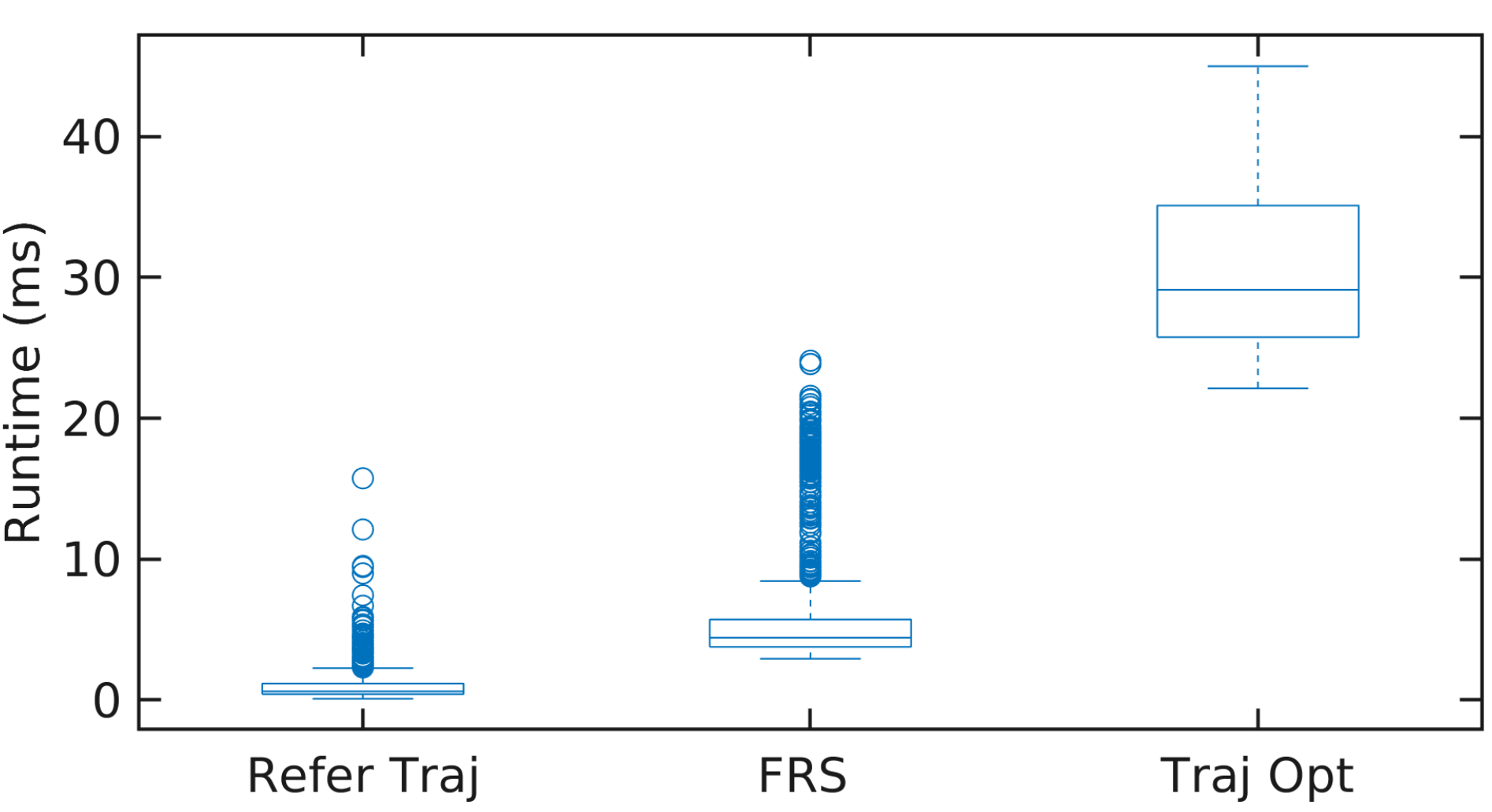}
	%	\hspace{1in}
	\caption{The breakdown of the computation time for the proposed motion planning framework.}
	\label{time_break}
	\vspace{-0.5cm}
\end{figure}

\subsection{Simulation Experiment}
\label{se}
\begin{figure*}[t]
	\vspace{0.2cm}
	\begin{center}
		\includegraphics[scale=0.145]{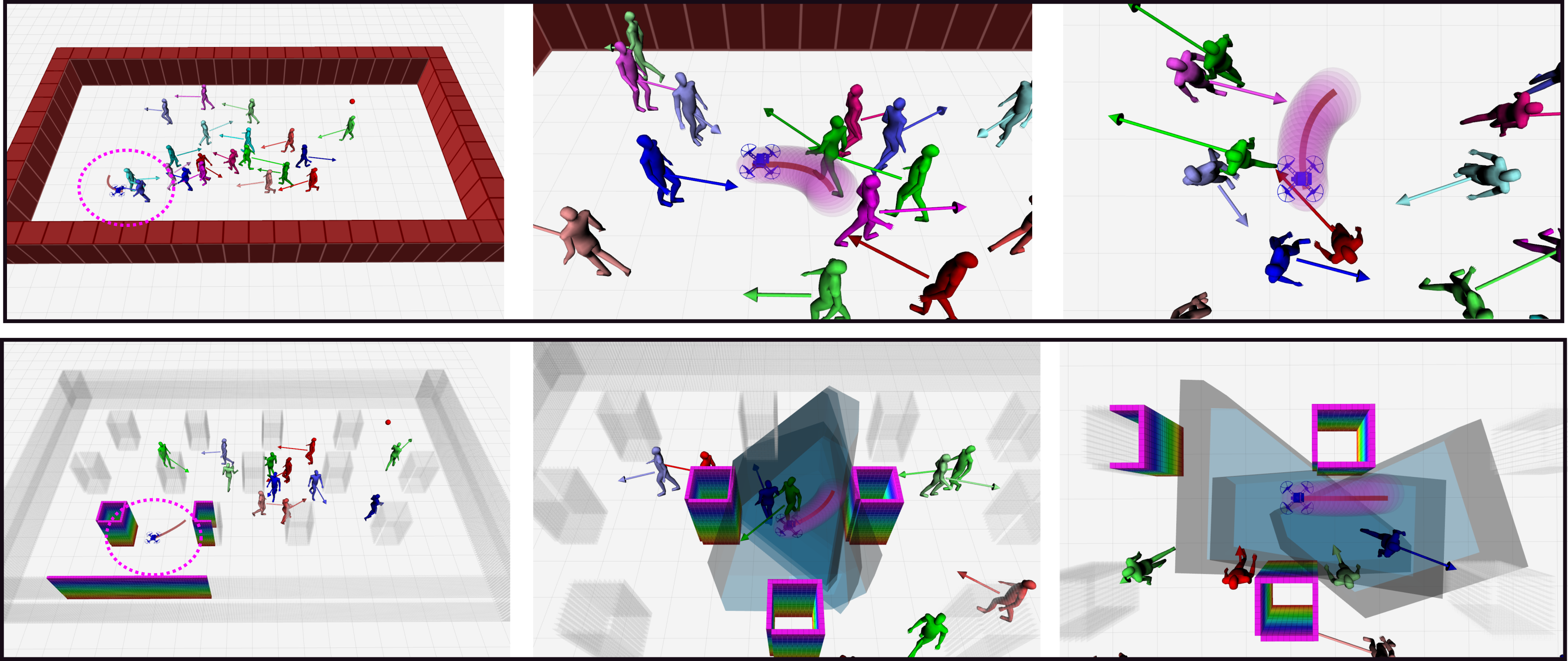}
	\end{center}
	\vspace{-0.5cm}
	\caption{Snapshots of our planning framework in simulation. \textbf{Top}: scenario \uppercase\expandafter{\romannumeral1}. \textbf{Bottom}: scenario \uppercase\expandafter{\romannumeral2}. \textbf{Arrow vector}: velocity of the human. \textbf{Colored map}: known static structure. \textbf{Translucent map}: unknown static structure. \textbf{Blue polyhedron}: flight corridor. Other labels are the same as \hyperref[headpic]{Figure \ref{headpic}}.}
	\label{sim_snap}	
	\vspace{-0.5cm}
\end{figure*}

The experiments are conducted in two scenarios: crowd navigation (CN) and factory patrolling (FP), as shown in \hyperref[sim_snap]{Figure \ref{sim_snap}}. In the first scenario, the robot navigates in a $ 20 \times 10 \, m $ environment populated with $ 20 $ humans. In the second scenario, the robot travels in a $ 20 \times 12 \, m $ environment with factory-like static structure and $ 12  $ randomly walking humans.
 All humans are driven by the social force model \footnote{https://github.com/srl-freiburg/pedsim\_ros}
  with randomly generated goals in the space and an update rate of $ 50\,Hz $. 
 The mission of robot is to move back and forth between the two corners on the diagonal.
  The noise on the position of robot and humans is set to $ \Sigma_p = \diag(0.05, 0.05, 0.05) \, m^2 $ and $ \Sigma_q = \diag(0.05, 0.05, 0) \, m^2 $, respectively. The velocity uncertainty is considered as $ \diag(0.03, 0.03, 0.03) \, m^2/s^2 $ and $ \Sigma_p = \diag(0.03, 0.03, 0) \, m^2/s^2 $. The shapes of robot and human are modeled as ellipsoids with semi-axis $ [0.22, 0.22, 0.1]\, m $ and $ [0.3, 0.3, 0.875]\, m $, respectively. 
   For our planning framework, we select a planning horizon of $ 1 \, s $ with $ 20 $ steps and the obstacles within a 5-meter radius of the robot are taken into consideration. The prescribed trajectory risk is $0.2$ with uniform allocation strategy $ \Delta_t = 0.2 / 20 = 0.01 $. The bounded disturbance on the external force is $ 0.3 \, mg $. The limit of the trajectory optimization time is $ 45 \, ms $ and the maximum speed of the robot and humans is $ 2 \, m/s $. We use the code generation engine FORCES Pro \cite{FORCESNLP} to generate our optimization solver and implement our motion planning framework in C++ 11. The simulations are run on a laptop with Intel i7-9750H CPU.

We compared our work with \cite{xu2022dpmpc} and \cite{wang2021autonomous}, which can handle the static environment with dynamic obstacles. Each method runs for two hours in the two scenarios and the average performance on success rate, mission completion time, trajectory risk and computation time is compared.
The planning is regarded as failure whenever the robot collides with moving humans. The trajectory and computation time is only calculated when there exist humans in the planning horizon. 
From the \hyperref[sim_bench]{Table \ref{sim_bench}}
,  we can conclude that the proposed planning framework is more reliable and less conservative. Compared to \cite{wang2021autonomous}, our method has a longer computation time but still meets the real-time planning requirement.
 
\begin{table}
	\centering
	\fontsize{8}{11} \selectfont
	\vspace{0.2cm}
	\caption{Comparison of Motion planning Framework } 
	\begin{tabular}{cc|ccccc}
		\toprule
		\toprule
		\multicolumn{2}{c|}{Scenarios}  &  Succ R (\%) & Mis T (s)  & Risk & Comp T (ms)  \cr
		
		\hline
		
		\multirow{3}{*}{CN} 
		& Ours & \textbf{96}  & \textbf{22.5} & \textbf{0.18} & 52.7   \cr
		&  \cite{xu2022dpmpc} & 90  & 26.2 & 0.016 & 43.4   \cr
		&  \cite{wang2021autonomous} & 78  & 27.1 & 3.5 & \textbf{8.3}  \cr
		
		\hline
		\multirow{3}{*}{FP} & Ours & \textbf{99}  & 15.7 & \textbf{0.15} & 45.5 \cr
		&  \cite{xu2022dpmpc} & 92  & 23.8 & 0.004 & 38.6 \cr
		&  \cite{wang2021autonomous} & 85  & \textbf{13.9} & 2.6 & \textbf{9.1} \cr
		
		\bottomrule
		\bottomrule		   
	\end{tabular}
	\label{sim_bench}
	\vspace{-0.5cm}
\end{table}

%Overall, this experiment provides a setting to test the robot's ability to navigate in highly uncertain environment, and demonstrates the effectiveness of the planning algorithm in achieving its goal.

\subsection{Real-world Experiment}
\setlength{\intextsep}{0pt}

As illustrated in \hyperref[headpic]{Figure \ref{headpic}}, the robot is faced with an unintentionally aggressive human walking in its vicinity, along with the wind disturbance from a nearby fan. The robot is ordered to hover in a fixed position and respond appropriately when the human is at risk of colliding with it. 
To track the movements of both the robot and the human, a motion capture system is employed to capture their poses, which are then processed by the EKF algorithm with the models (\ref{uav_model}) and (\ref{cvm}). 
The human's position and velocity uncertainty are set to $ \Sigma_q = \diag(0.10, 0.10, 0) , m^2 $ and $ \Sigma_q = \diag(0.10, 0.10, 0) , m^2/s^2 $, respectively. The allocated collision probability for each step is set to the value of $0.03$. Furthermore, to account for the impact of wind, an external force of $1.5 \, mg$ is set in the planning parameters.
All the other parameters are the same as those described in Section \ref{se}. The MPC is running on the onboard computer Orin NX, and the breakdown of the computation time is displayed in \hyperref[time_break]{Figure \ref{time_break}}. As shown in the accompanying video,  our planner can reliably operate in highly uncertain environments with real-time performance.

\setlength{\intextsep}{0pt}

\section{Conclusion}
\label{cln}
This paper proposed a reliable motion planning framework for UAVs operating in highly uncertain environments with dynamic obstacles and external disturbances. To enhance the efficiency of the planned trajectory, we proposed a tight upper bound for collision probability and evaluated it using both exact and approximate methods. Future research directions include integrating our visibility planning algorithm \cite{liu2022star} to reduce localization uncertainty and developing a robust planning framework that can effectively handle the challenges of complex environments.

\addtolength{\textheight}{0cm}   % This command serves to balance the column lengths

%%%%%%%%%%%%%%%%%%%%%%%%%%%%%%%%%%%%%%%%%%%%%%%%%%%%%%%%%%%%%%%%%%%%%%%%%%%%%%%%

%%%%%%%%%%%%%%%%%%%%%%%%%%%%%%%%%%%%%%%%%%%%%%%%%%%%%%%%%%%%%%%%%%%%%%%%%%%%%%%%

%%%%%%%%%%%%%%%%%%%%%%%%%%%%%%%%%%%%%%%%%%%%%%%%%%%%%%%%%%%%%%%%%%%%%%%%%%%%%%%%
%\section*{APPENDIX}

%\section*{ACKNOWLEDGMENT}

%The preferred spelling of the word `acknowledgment' in America is without an `e' after the `g'. Avoid the stilted expression, `One of us (R. B. G.) thanks . . .'  Instead, try `R. B. G. thanks'. Put sponsor acknowledgments in the unnumbered footnote on the first page.

%%%%%%%%%%%%%%%%%%%%%%%%%%%%%%%%%%%%%%%%%%%%%%%%%%%%%%%%%%%%%%%%%%%%%%%%%%%%%%%%

%References are important to the reader; therefore, each citation must be complete and correct. If at all possible, references should be commonly available publications.
%They are cited like so: \cite{IEEEexample:articleetal}, \cite{IEEEexample:book}, ...
%\newpage

\bibliographystyle{IEEEtran}
\bibliography{IEEEabrv,test}

\begin{thebibliography}{10}
\providecommand{\url}[1]{#1}
\csname url@samestyle\endcsname
\providecommand{\newblock}{\relax}
\providecommand{\bibinfo}[2]{#2}
\providecommand{\BIBentrySTDinterwordspacing}{\spaceskip=0pt\relax}
\providecommand{\BIBentryALTinterwordstretchfactor}{4}
\providecommand{\BIBentryALTinterwordspacing}{\spaceskip=\fontdimen2\font plus
\BIBentryALTinterwordstretchfactor\fontdimen3\font minus
  \fontdimen4\font\relax}
\providecommand{\BIBforeignlanguage}[2]{{%
\expandafter\ifx\csname l@#1\endcsname\relax
\typeout{** WARNING: IEEEtran.bst: No hyphenation pattern has been}%
\typeout{** loaded for the language `#1'. Using the pattern for}%
\typeout{** the default language instead.}%
\else
\language=\csname l@#1\endcsname
\fi
#2}}
\providecommand{\BIBdecl}{\relax}
\BIBdecl

\bibitem{blackmore2011chance}
L.~Blackmore, M.~Ono, and B.~C. Williams, ``Chance-constrained optimal path
  planning with obstacles,'' \emph{IEEE Transactions on Robotics}, vol.~27,
  no.~6, pp. 1080--1094, 2011.

\bibitem{majumdar2017funnel}
A.~Majumdar and R.~Tedrake, ``Funnel libraries for real-time robust feedback
  motion planning,'' \emph{The International Journal of Robotics Research},
  vol.~36, no.~8, pp. 947--982, 2017.

\bibitem{zhu2019chance}
H.~Zhu and J.~Alonso-Mora, ``Chance-constrained collision avoidance for mavs in
  dynamic environments,'' \emph{IEEE Robotics and Automation Letters}, vol.~4,
  no.~2, pp. 776--783, 2019.

\bibitem{lin2020robust}
J.~Lin, H.~Zhu, and J.~Alonso-Mora, ``Robust vision-based obstacle avoidance
  for micro aerial vehicles in dynamic environments,'' in \emph{2020 IEEE
  International Conference on Robotics and Automation (ICRA)}.\hskip 1em plus
  0.5em minus 0.4em\relax IEEE, 2020, pp. 2682--2688.

\bibitem{castillo2020real}
M.~Castillo-Lopez, P.~Ludivig, S.~A. Sajadi-Alamdari, J.~L. Sanchez-Lopez,
  M.~A. Olivares-Mendez, and H.~Voos, ``A real-time approach for
  chance-constrained motion planning with dynamic obstacles,'' \emph{IEEE
  Robotics and Automation Letters}, vol.~5, no.~2, pp. 3620--3625, 2020.

\bibitem{dai2019chance}
S.~Dai, S.~Schaffert, A.~Jasour, A.~Hofmann, and B.~Williams, ``Chance
  constrained motion planning for high-dimensional robots,'' in \emph{2019
  International Conference on Robotics and Automation (ICRA)}.\hskip 1em plus
  0.5em minus 0.4em\relax IEEE, 2019, pp. 8805--8811.

\bibitem{manchester2017dirtrel}
Z.~Manchester and S.~Kuindersma, ``Dirtrel: Robust trajectory optimization with
  ellipsoidal disturbances and lqr feedback.'' in \emph{Robotics: Science and
  Systems}.\hskip 1em plus 0.5em minus 0.4em\relax Cambridge, MA, USA, 2017.

\bibitem{seo2019robust}
H.~Seo, D.~Lee, C.~Y. Son, C.~J. Tomlin, and H.~J. Kim, ``Robust trajectory
  planning for a multirotor against disturbance based on hamilton-jacobi
  reachability analysis,'' in \emph{2019 IEEE/RSJ International Conference on
  Intelligent Robots and Systems (IROS)}.\hskip 1em plus 0.5em minus
  0.4em\relax IEEE, 2019, pp. 3150--3157.

\bibitem{lambert2008fast}
A.~Lambert, D.~Gruyer, and G.~Saint~Pierre, ``A fast monte carlo algorithm for
  collision probability estimation,'' in \emph{2008 10th International
  Conference on Control, Automation, Robotics and Vision}.\hskip 1em plus 0.5em
  minus 0.4em\relax IEEE, 2008, pp. 406--411.

\bibitem{park2012itomp}
C.~Park, J.~Pan, and D.~Manocha, ``Itomp: Incremental trajectory optimization
  for real-time replanning in dynamic environments,'' in \emph{Twenty-Second
  International Conference on Automated Planning and Scheduling}, 2012.

\bibitem{du2011probabilistic}
N.~E. Du~Toit and J.~W. Burdick, ``Probabilistic collision checking with chance
  constraints,'' \emph{IEEE Transactions on Robotics}, vol.~27, no.~4, pp.
  809--815, 2011.

\bibitem{hardy2013contingency}
J.~Hardy and M.~Campbell, ``Contingency planning over probabilistic obstacle
  predictions for autonomous road vehicles,'' \emph{IEEE Transactions on
  Robotics}, vol.~29, no.~4, pp. 913--929, 2013.

\bibitem{thomas2021exact}
A.~Thomas, F.~Mastrogiovanni, and M.~Baglietto, ``Exact and bounded collision
  probability for motion planning under gaussian uncertainty,'' \emph{IEEE
  Robotics and Automation Letters}, vol.~7, no.~1, pp. 167--174, 2021.

\bibitem{wu2021external}
Y.~Wu, Z.~Ding, C.~Xu, and F.~Gao, ``External forces resilient safe motion
  planning for quadrotor,'' \emph{IEEE Robotics and Automation Letters},
  vol.~6, no.~4, pp. 8506--8513, 2021.

\bibitem{kim2018computing}
S.~Kim, D.~Falanga, and D.~Scaramuzza, ``Computing the forward reachable set
  for a multirotor under first-order aerodynamic effects,'' \emph{IEEE Robotics
  and Automation Letters}, vol.~3, no.~4, pp. 2934--2941, 2018.

\bibitem{kamel2017robust}
M.~Kamel, J.~Alonso-Mora, R.~Siegwart, and J.~Nieto, ``Robust collision
  avoidance for multiple micro aerial vehicles using nonlinear model predictive
  control,'' in \emph{2017 IEEE/RSJ International Conference on Intelligent
  Robots and Systems (IROS)}.\hskip 1em plus 0.5em minus 0.4em\relax IEEE,
  2017, pp. 236--243.

\bibitem{wang2021autonomous}
Y.~Wang, J.~Ji, Q.~Wang, C.~Xu, and F.~Gao, ``Autonomous flights in dynamic
  environments with onboard vision,'' in \emph{2021 IEEE/RSJ International
  Conference on Intelligent Robots and Systems (IROS)}.\hskip 1em plus 0.5em
  minus 0.4em\relax IEEE, 2021, pp. 1966--1973.

\bibitem{ho2021adaptive}
C.~Ho, J.~Patrikar, R.~Bonatti, and S.~Scherer, ``Adaptive safety margin
  estimation for safe real-time replanning under time-varying disturbance,''
  \emph{arXiv preprint arXiv:2110.03119}, 2021.

\bibitem{rimon1997obstacle}
E.~Rimon and S.~P. Boyd, ``Obstacle collision detection using best ellipsoid
  fit,'' \emph{Journal of Intelligent and Robotic Systems}, vol.~18, pp.
  105--126, 1997.

\bibitem{0Quadratic}
A.~M. Mathai and S.~B. Provost, \emph{Quadratic Forms in Random Variables:
  Theory and Applications}.\hskip 1em plus 0.5em minus 0.4em\relax Quadratic
  forms in random variables :.

\bibitem{thomas2021integrated}
A.~Thomas, F.~Mastrogiovanni, and M.~Baglietto, ``An integrated localization,
  motion planning and obstacle avoidance algorithm in belief space,''
  \emph{Intelligent Service Robotics}, vol.~14, pp. 235--250, 2021.

\bibitem{kotz1967series}
S.~Kotz, N.~L. Johnson, and D.~Boyd, ``Series representations of distributions
  of quadratic forms in normal variables. i. central case,'' \emph{The Annals
  of Mathematical Statistics}, vol.~38, no.~3, pp. 823--837, 1967.

\bibitem{frey2020collision}
K.~M. Frey, T.~J. Steiner, and J.~P. How, ``Collision probabilities for
  continuous-time systems without sampling [with appendices],'' \emph{arXiv
  preprint arXiv:2006.01109}, 2020.

\bibitem{park2018fast}
C.~Park, J.~S. Park, and D.~Manocha, ``Fast and bounded probabilistic collision
  detection for high-dof trajectory planning in dynamic environments,''
  \emph{IEEE Transactions on Automation Science and Engineering}, vol.~15,
  no.~3, pp. 980--991, 2018.

\bibitem{Andersson2018}
J.~A.~E. Andersson, J.~Gillis, G.~Horn, J.~B. Rawlings, and M.~Diehl,
  ``{CasADi} -- {A} software framework for nonlinear optimization and optimal
  control,'' \emph{Mathematical Programming Computation}, In Press, 2018.

\bibitem{scholler2020constant}
C.~Sch{\"o}ller, V.~Aravantinos, F.~Lay, and A.~Knoll, ``What the constant
  velocity model can teach us about pedestrian motion prediction,'' \emph{IEEE
  Robotics and Automation Letters}, vol.~5, no.~2, pp. 1696--1703, 2020.

\bibitem{zhou2019robust}
B.~Zhou, F.~Gao, L.~Wang, C.~Liu, and S.~Shen, ``Robust and efficient quadrotor
  trajectory generation for fast autonomous flight,'' \emph{IEEE Robotics and
  Automation Letters}, vol.~4, no.~4, pp. 3529--3536, 2019.

\bibitem{liu2017planning}
S.~Liu, M.~Watterson, K.~Mohta, K.~Sun, S.~Bhattacharya, C.~J. Taylor, and
  V.~Kumar, ``Planning dynamically feasible trajectories for quadrotors using
  safe flight corridors in 3-d complex environments,'' \emph{IEEE Robotics and
  Automation Letters}, vol.~2, no.~3, pp. 1688--1695, 2017.

\bibitem{FORCESNLP}
A.~Zanelli, A.~Domahidi, J.~Jerez, and M.~Morari, ``Forces nlp: an efficient
  implementation of interior-point... methods for multistage nonlinear
  nonconvex programs,'' \emph{International Journal of Control}, pp. 1--17,
  2017.

\bibitem{xu2022dpmpc}
Z.~Xu, D.~Deng, Y.~Dong, and K.~Shimada, ``Dpmpc-planner: A real-time uav
  trajectory planning framework for complex static environments with dynamic
  obstacles,'' in \emph{2022 International Conference on Robotics and
  Automation (ICRA)}.\hskip 1em plus 0.5em minus 0.4em\relax IEEE, 2022, pp.
  250--256.

\bibitem{liu2022star}
T.~Liu, Q.~Wang, X.~Zhong, Z.~Wang, C.~Xu, F.~Zhang, and F.~Gao, ``Star-convex
  constrained optimization for visibility planning with application to aerial
  inspection,'' in \emph{2022 International Conference on Robotics and
  Automation (ICRA)}.\hskip 1em plus 0.5em minus 0.4em\relax IEEE, 2022, pp.
  7861--7867.

\end{thebibliography}

\end{document}